\documentclass[conference]{IEEEtran}

\ifCLASSINFOpdf
\usepackage{graphicx}
\else
\fi
\usepackage[caption=false,font=footnotesize]{subfig}

\usepackage{graphicx}
\usepackage[noend]{algorithmic}
\usepackage{algorithm}
\usepackage{array}
\usepackage{amsmath}

\begin{document}
%
\title{Technical Report CMP-C14-01: An Experimental Evaluation of Nearest Neighbour Time Series Classification}

\author{\IEEEauthorblockN{Anthony Bagnall}
\IEEEauthorblockA{School of Computing Sciences\\
University of East Anglia\\
Norwich, UK
ajb@uea.ac.uk}
\and
\IEEEauthorblockN{Jason Lines}
\IEEEauthorblockA{School of Computing Sciences\\
University of East Anglia\\
Norwich, UK
jason.lines@uea.ac.uk}
}

\maketitle

\begin{abstract}
Data mining research into time series classification (TSC) has focussed on alternative distance measures for nearest neighbour classifiers. It is standard practice to use 1-NN with Euclidean or dynamic time warping (DTW) distance as a straw man for comparison.
As part of a wider investigation into elastic distance measures for TSC~\cite{lines14elastic}, we perform a series of experiments to test whether this standard practice is valid.

Specifically, we compare 1-NN classifiers with Euclidean and DTW distance to standard classifiers, examine whether the performance of 1-NN Euclidean approaches that of 1-NN DTW as the number of cases increases,  assess whether there is any benefit of setting $k$ for $k$-NN through cross validation whether it is worth setting the warping path for DTW through cross validation  and finally is it better to use a window or weighting for DTW. Based on experiments on 77 problems, we conclude that 1-NN with Euclidean distance is fairly easy to beat but 1-NN with DTW is not, if window size is set through cross validation.

\end{abstract}

\section{Introduction}

Time-series classification (TSC) problems involve training a classifier on a set of cases, where each case contains an ordered set of real valued attributes and a class label. Time-series classification problems arise in a wide range of  fields including, but not limited to, data mining, statistics, machine learning, signal processing, environmental sciences, computational biology, image processing and chemometrics. A wide range of algorithms have been proposed for solving TSC problems (see, for example~\cite{rodriguez05svm,ding08querying,corduas08timeseries,ye09shapelets,abraham10integrated,batista11complexity,jeong11weighted,bagnall12ensemble,douzal12trees,hills13shapelet,deng13forest}).

In~\cite{bagnall12ensemble}, Bagnall {\em et al.} argue that the easiest way to gain improvement in accuracy on TSC problems is to transform into an alternative data space where the discriminatory features are more easily detected. They constructed classifiers on data in the time, frequency, autocorrelation  and principal component domains and combined predictions through alternative ensemble schemes. The main conclusion in~\cite{bagnall12ensemble} is that for problems where the discriminatory features are based on  similarity in change and  similarity in shape, operating in a different data space produces a better performance improvement than designing a more complex classifier for the time domain. However, the issue of what is the best technique in a single data domain is not addressed in~\cite{bagnall12ensemble}. Our aim is to experimentally determine the best method for constructing classifiers in the time domain, an area that has drawn most of the attention of TSC data mining researchers~\cite{batista11complexity,jeong11weighted,rodriguez05svm,douzal12trees,deng13forest}. The general consensus amongst data mining researchers is that {\em ``simple nearest neighbor classification is very difficult to beat"}~\cite{batista11complexity}. For problems with few training cases, an elastic distance measure such as dynamic time warping (DTW) or longest common subsequence (LCSS) is often superior to Euclidean distance, but as the number of series increases {\em``the accuracy of elastic measures converge with that of Euclidean distance"}~\cite{ding08querying}.

Our  objective is to empirically test various aspects of these commonly made assertions. We examine whether one nearest neighbour (1-NN) is in fact not significantly worse than other classifiers and whether setting the parameter $k$ for nearest neighbour through cross validation on the training data improves performance. For DTW, the key parameter is the warping window size. This dictates the largest allowable displacement between two points in the warping path.
We evaluate whether setting the warping window through cross validation makes the classifier more accurate. Several alternative forms of DTW have been proposed in the literature. A version of DTW that weights against large warpings (WDTW) is described in~\cite{jeong11weighted}. The weighting scheme can be used in conjunction with dynamic time warping and an alternative version based on first order differences (DDTW), described in~\cite{keogh01derivative}. We extend the experiments described in~\cite{jeong11weighted} to test whether their conclusions hold over a large number of data sets with parameter optimisations for all of the algorithms considered. All datasets and code to reproduce experiments and results are available online~\cite{TSC_Web}.

To summarise, we have conducted extensive experiments to answer the following questions:

 \begin{enumerate}
 \item Is Euclidean/DTW nearest neighbour (1-NN) really better than other commonly used algorithms such as tree-based or probabilistic classifiers?
     \item Does the accuracy of 1-NN Euclidean approach that of 1-NN DTW as training set size increases?
 \item Is it better to use $k$ nearest neighbours ($k$-NN), with $k$ set through cross validation, rather than 1-NN?
 \item Is it worthwhile setting the warping window for DTW through cross validation?
 \end{enumerate}

 We have answered these questions through over three million experiments on 77 TSC problems. 43 datasets come from the the UCR repository~\cite{UCRWeb}, 24 problems are from other published research, including~\cite{ye09shapelets,ShapeletWeb}, and 5 are new data sets on electricity device classification problems described in~\cite{lines14elastic}.

The structure of this paper is as follows. In Section~\ref{background} we provide background into TSC and review related research on DTW. In Section~\ref{data} we detail the 77 data sets we used in experiments.  In Section~\ref{results} we present our results and in Section~\ref{conc} we summarise our conclusions.

\section{Background and Related Work}
\label{background}

\subsection{Time Series Classification (TSC)}

We define time series classification as the problem of building a classifier from a collection of labelled training time series. We limit our attention to problems where each time series has the same number of observations. We define a time series $\bf{x_i}$ as a set of ordered observations
$${\bf x_i} = <x_{i1},\ldots,x_{im}>$$ and an associated class label $y_i$. The training set is a set of $n$ labelled pairs $$D=\{({\bf x_1}, y_1), \ldots, ({\bf x_n}, y_n)\}.$$  For traditional classification problems, the order of the attributes is unimportant and the interaction between variables is considered independent of their relative positions. For time series data, the ordering of the variables is often crucial in finding the best discriminating features. There are three broad categories of TSC discriminating features which are described by three general approaches to measuring similarity between time series: similarity in shape, similarity in change and similarity in time.

Similarity in shape describes the scenario where class membership is characterised by a common shape but the discriminatory shape is phase independent. If the common shape involves the whole series, but is  phase shifted between instances of the same class, then transformation into the frequency domain is often the best approach (for example, see~\cite{agrawal93efficient}). If the common shape is local and embedded in confounding noise, then subsequence techniques such as Shapelets can be employed~\cite{ye09shapelets,hills13shapelet}.

Similarity in change refers to the situation where the relevant discriminatory features are related to the autocorrelation function of each series. The most common approach in this situation is to fit an ARMA model, then base similarity on differences in model parameters~\cite{corduas08timeseries}. The common element to similarity in shape and change is that similarity between series is not measured in the time domain.

However, the majority of the data mining research into TSC has concentrated in similarity in time. This can be quantified by measures such as Euclidean distance or correlation~\cite{douzal12trees,abraham10integrated}. Similarity in time is characterised by the situation where the series from each class are observations of an underlying common curve in the time dimension. Variation around this underlying common shape is caused by noise in observation, and also by possible noise in indexing which may cause a slight phase shift. A classic example of this type of similarity is the Cylinder-Bell-Funnel artificial data set, where there is noise around the underlying shape, but also noise in the index of where the underlying shape transitions~\cite{douzal12trees}.

The commonly used benchmark classification algorithm for problems with small phase shift is 1-NN with an elastic measure such as DTW or LCSS to allow for small shifts in the time axis. In a comprehensive study~\cite{ding08querying}, DTW was found to be as least as good as other elastic measures based on edit distance, and constraining the warping window was found to speed up computation, {\em ``while yielding the same or even better accuracy"}~\cite{ding08querying}. The experimentation in~\cite{ding08querying} addresses the issue of what distance measure to use and is the starting point for our research. We begin by testing their assumptions about classifier selection and parameter setting before investigating alternative variants of DTW and combination schemes for the classifiers.

\subsection{Classification Algorithms}

 The nearest neighbour classifier is a lazy classifier (i.e. requires no training) that classifies new cases by finding the closest case in the training set with a distance function, then using the class of the closest case as the predicted class for the new case. Given the focus on distance functions in time series data mining research, it is perhaps unsurprising that the majority of classification has used 1-NN. Whilst often highly effective, 1-NN is known to be susceptible to problems such as outliers in the training set and large numbers of redundant features. Outliers can be compensated for by using $k$ nearest neighbours and a voting scheme. Redundant features may be dealt with by filtering or by employing one of the plethora of alternative classifiers. Filtering was found to be not effective in~\cite{bagnall12ensemble}. We compare nearest neighbour classifiers against C4.5~\cite{quinlan93c4}, Random Forest~\cite{breiman01random}, Rotation Forest~\cite{rodriguez06rotation},  Naive Bayes~\cite{lewis98naive} and Bayesian networks~\cite{pearl88bayesnet} and Support Vector Machines with linear and quadratic kernels~\cite{cortes95support}.

\subsection{Dynamic Time Warping}

For similarity in shape, Dynamic Time Warping (DTW) is commonly used to mitigate against distortions in the time
axis~\cite{ratanamahatana05threemyths}. Suppose we want to measure the distance between two series, $\mathbf{a}=\{a_1,a_2,\ldots,a_m\}$ and $\mathbf{b}=\{b_1,b_2,\ldots,b_m\}$. Let $M(\mathbf{a},\mathbf{b})$ be the $m \times
m$ pointwise distance matrix between $\mathbf{a}$ and $\mathbf{b}$, where
$M_{i,j}=   (a_i-b_j)^2$.

A warping path $P=<(e_1,f_1),(e_2,f_2),\ldots,(e_s,f_s)>$ is a set of points (i.e. pairs of indexes) that
define a traversal of matrix $M$. So, for example, the Euclidean distance $d_E(\mathbf{a,b})=\sum_{i=1}^m (a_i-b_i)^2$ is the path along the diagonal of $M$, i.e.

$P_e=<(1,1,(2,2),\ldots,(m,m)>$.

A valid warping path must satisfy the conditions $(e_1,f_1)=(1,1)$ and $(e_s,f_s)=(m,m)$ and
that $0 \leq e_{i+1}-e_{i} \leq 1$ and $0 \leq f_{i+1}- f_i \leq 1$
for all $i < m$.

The DTW distance between series is the path through $M$ that minimizes the total distance, subject to constraints on
the amount of warping allowed. Let $p_i=M_{a_{e_i},b_{f_i}}$ be the distance
between element at position $e_i$ of $\mathbf{a}$ and at position $f_i$ of $\mathbf{b}$ for the $i^{th}$
pair of points in a proposed warping path $P$. The distance for any path $P$ is

\[ D_P(\mathbf{a},\mathbf{b}) =\sum_{i=1}^s p_i.\]

If $\mathcal{P}$ is the space of all possible paths, the DTW path $P^*$ is the path that has the minimum distance, i.e.
$$P^* = \min_{P \in \mathcal{P}}(D_P(\mathbf{a},\mathbf{b})),$$
and hence the DTW distance between series is
$$D_{P*}(\mathbf{a},\mathbf{b}) = \sum_{i=1}^k p_i.$$

The optimal path $P^*$ can be found exactly through a dynamic programming formulation. This can be a time consuming operation, and it is common to put a restriction on the amount of warping allowed. This restriction is equivalent to putting a maximum allowable distance between any pairs of indexes in a proposed path. If the warping window,  $r$, is the proportion of warping allowed, then the optimal path is constrained so that $$|e_i-f_i| \leq r\cdot m \;\;\; \forall (e_i,f_i) \in P^*.$$

\subsection{Longest Common Subsequence}

\begin{figure}[htbp]
	\centering
       \includegraphics[width=4cm]{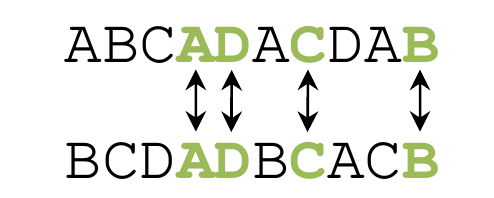}
	\caption{The point-wise matches for ``ABCADACDAB" and ``BCDADBCACB"}
	\label{img:lcssPointwise}
\end{figure}

The Longest Common Subsequence Distance (LCSS) is based on the solution to the longest common subsequence problem in pattern matching. The typical problem is to find the longest subsequence that is common to two discrete series based on the edit distance. This approach can be extended to consider real-valued time series by using a distance threshold $\epsilon$, which defines the maximum difference between a pair of values that is allowed for them to be considered a match~\cite{kuzmanic07handshape}.  LCSS finds the optimal alignment between two series by inserting gaps to find the greatest number of matching pairs. For example, consider the discrete case where we have two strings S = ``ABCADACDAB" and T = ``BCDADBCACB". If we simply observe the point-wise matches between the two sequences, we can extract matching pairs for the substring ``ADCB" as shown in Figure~\ref{img:lcssPointwise}.

\begin{figure}[htbp]
	\centering
       \includegraphics[width=4cm]{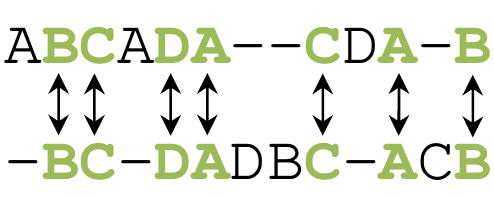}
	\caption{The LCSS of ``ABCADACDAB" and ``BCDADBCACB"}
	\label{img:lcssMatchedExample}
\end{figure}

However, using LCSS we can find a longer matching subsequence by inserting spaces into the two strings. In this case, the elasticity of the measure means we find the subsequence ``BCDACAB", as depicted in Figure~\ref{img:lcssMatchedExample}.

The LCSS between two series ${\bf a}$ and ${\bf b}$ can be found using Algorithm~\ref{algo1}.

\begin{algorithm}[htbp]
	\caption{LCSS (${\bf a},{\bf b}$)}
\label{algo1}
	\begin{algorithmic}[1]
\STATE Let $L$ be an $(m+1)\times(m+1)$ matrix initialised to zero.
\FOR{$i \leftarrow  m$ to $1$}
    \FOR{$j \leftarrow  m$ to $1$}
        \STATE $L_{i,j} \leftarrow L_{i+1,j+1}$
        \IF{$a_i = b_j$}
            \STATE $L_{i,j} \leftarrow L_{i,j}+1$
        \ELSIF{$L_{i,j+1} > L_{i,j}$}
            \STATE $L_{i,j} \leftarrow L_{i,j+1}$
        \ELSIF {$L_{i+1,j} > L_{i,j}$}
            \STATE $L_{i,j} \leftarrow L_{i+1,j}$
        \ENDIF
   \ENDFOR
\ENDFOR
\RETURN $L_{1,1}$
	\end{algorithmic}
\end{algorithm}

The LCSS distance between ${\bf a}$ and ${\bf b}$ is

\[d_{LCSS}({\bf a,b}) = 1- \frac{LCSS({\bf a,b})}{m}.\]

\subsection{Derivative Dynamic Time Warping}

Keogh and Pazzani proposed a modification of DTW called Derivative Dynamic Time Warping (DDTW)~\cite{keogh01derivative} that first transforms the series into a series of first differences. Given a series $\mathbf{a}=\{a_1,a_2,\ldots,a_m\}$, the difference series is $\mathbf{a'}=\{a'_2,a'_2,\ldots,a'_{m-1}\}$ where $a'_i$ is defined as the average of the slopes between $a_{i-1}$ and $a_i$ and $a_i$ and $a_{i+1}$, i.e.

$$a'_i = \frac{(a_i-a_{i-1})+(a_{i+1}-a_{i-1})/2}{2},$$

for $1<i<m$. DDTW is designed to mitigate against noise in the series that can adversely affect DTW.

\subsection{Weighted Dynamic Time Warping}

A weighted form of DTW (WDTW) was proposed by Jeong {\em et al.}~\cite{jeong11weighted}. WDTW adds a multiplicative weight penalty based on the warping distance between points in the warping path. It favours reduced warping, and is a smooth alternative to the cutoff point approach of using a warping window. When creating the distance matrix $M$, a weight penalty  $w_{|i-j|}$ for a warping distance of  $|i-j|$ is applied, so that

$$M_{i,j}=  w_{|i-j|} (a_i-b_j)^2.$$

A logistic weight function is proposed in~\cite{jeong11weighted}, so that a warping of $a$ places imposes a weighting of

$$w(a)=\frac{w_{max}}{1+e^{-g\cdot(a-m/2)}},$$
where $w_{max}$ is an upper bound on the weight (set to 1), $m$ is the series length and $g$ is a parameter that controls the penalty level for large warpings. The larger $g$ is, the greater the penalty for warping.

Jeong {\em et al.} compared WDTW to Euclidean distance, full window DTW and LCSS on 20 UCR data sets using a 1-NN classifier. Half of the test data was used as a validation set for setting the value of $g$ and the other half was used to measure accuracy. They state their results demonstrate {\em ``WDTW and WDDTW clearly outperform standard DTW, DDTW and LCSS measures."} We test this assertion in Section~\ref{results}.

\section{Data Sets}
\label{data}

We have collected 77 data sets, the names of which are shown in Table~\ref{tab1}. 43 of these are available from the UCR repository~\cite{UCRWeb}, 29 were used in other published work~\cite{hills13shapelet,ye09shapelets,ShapeletWeb} and 5 are new data sets we present for the first time. Further information and the data sets we have permission to circulate are available from~\cite{TSC_Web}.

\begin{table*}
\begin{center}
\caption{Data sets grouped by problem type. The actual file names are in a string array in class TimeSeriesClassification.fileNames}
\label{tab1}
\small
\begin{tabular}{|cccccc|} \hline
\multicolumn{6}{|c|}{\bf Image Outline Classification}\\ \hline
DistPhalanxAge  &DistPhalanxOutline &DistPhalanxTW  &FaceAll    &FaceFour  &WordSynonyms         \\
MidPhalanxAge   &MidPhalanxOutline  &MidPhalanxTW   &OSULeaf    &Phalanges  &yoga            \\
ProxPhalanxAge  &ProxPhalanxOutline &ProxPhalanxTW  &ShapesAll  &SwedishLeaf&MedicalImages \\
Symbols & Adiac   &ArrowHead      &BeetleFly          &BirdChicken    &DiatomSize\\
FacesUCR&fiftywords     &fish&HandOutlines  &Herring        & \\\hline
\multicolumn{6}{|c|}{\bf Motion Classification}\\ \hline

CricketX&CricketY&CricketZ&UWaveX&UWaveY&UWaveZ\\
UWaveAll&GunPoint&Haptics&InlineSkate & ToeSeg1&ToeSeg2\\
 \hline
\multicolumn{6}{|c|}{\bf Sensor Reading Classification}\\ \hline
Beef&Car&Chlorine&CinCECG&Coffee&Computers\\
FordA&FordB&ItalyPower&LargeKitchen&Lighting2&Lighting7\\
StarLightCurves&Trace&TwoLeadECG&wafer & RefrigerationDevices& MoteStrain\\
Earthquakes&ECG200&ECGFiveDays&ElectricDevices&SonyRobot1&SonyRobot2\\
OliveOil&Plane&ScreenType&SmallKitchenAppliances& MALLAT&\\
&&ECGThorax1&ECGThorax2&& \\ \hline
\multicolumn{6}{|c|}{\bf Simulated Classification Problems}\\ \hline
ARSim&CBF&SyntheticControl&ShapeletSim&TwoPatterns&\\ \hline

\end{tabular}
\end{center}
\end{table*}

We have grouped the problems into categories to help aid interpretation. The group of Sensor readings forms the largest category with 31 data sets. If we had more data sets it would be sensible to split the sensor categories into subtypes, such as human sensors and spectrographs. However, at this point such sub-typing would lead to groups that are too small. Image outline classification is the second largest category, with 29 problem sets. Many of the image outline problems, such as {\em BeetleFly}, are not rotationally aligned, and the expectation would be that classifiers in the time domain will not necessarily perform well with these data. The group of 14 Motion problems contains data taken from motion capture devices attached to human subjects. The final category is simulated data sets.

\section{Results}
\label{results}

 We conducted our classification experiments with WEKA~\cite{hall09weka} source code adapted for time series classification, using the 77 data sets described in Section~\ref{data}.

 All datasets are split into a training and testing set, and all parameter optimisation is conducted on the training set only. We use a train/test split for several reasons. Firstly, it is common practice to do so with the UCR datasets. Secondly, some of the data sets are designed so the train/test split removes bias. Combining to perform a cross validation would reintroduce the bias. And finally, it is not computationally feasible to cross validate everything. We ran over 3 million experiments on a 4148 core High Performance Cluster (with a theoretical peak performance of 65TFlops) over a period of a month. Adding another level of cross validation would have increased the time required, the number of experiments and the complexity of the code by an order of magnitude.

 For hypothesis testing purposes, we are assuming that these data sets are a random sample from the set of all possible TSC problems. This is true in the sense that we have not collected these data with any agenda in mind. We are not attempting to find data sets more suitable for one particular algorithm. However, as Table~\ref{tab1} demonstrates, there is a bias in areas of application we are considering. There are, for example, no problems from econometrics or finance. Because of this, we also present results split by problem type where appropriate. All the results are available on an Excel spreadsheet from~\cite{TSC_Web}.

\subsection{Are 1-NN Classifiers are hard to beat?}

Our null hypothesis is that the average accuracy of 1-NN Euclidean and 1-NN DTW are the same as other classifiers in the time domain. If we can reject this hypothesis for the one sided alternative that the average accuracy is significantly worse than at least one other classifier, then we have evidence that the  assertion {\em ''1-NN Classifiers are hard to beat"} is incorrect.

\begin{figure}[*h]
	\centering
        \includegraphics[width=9cm]{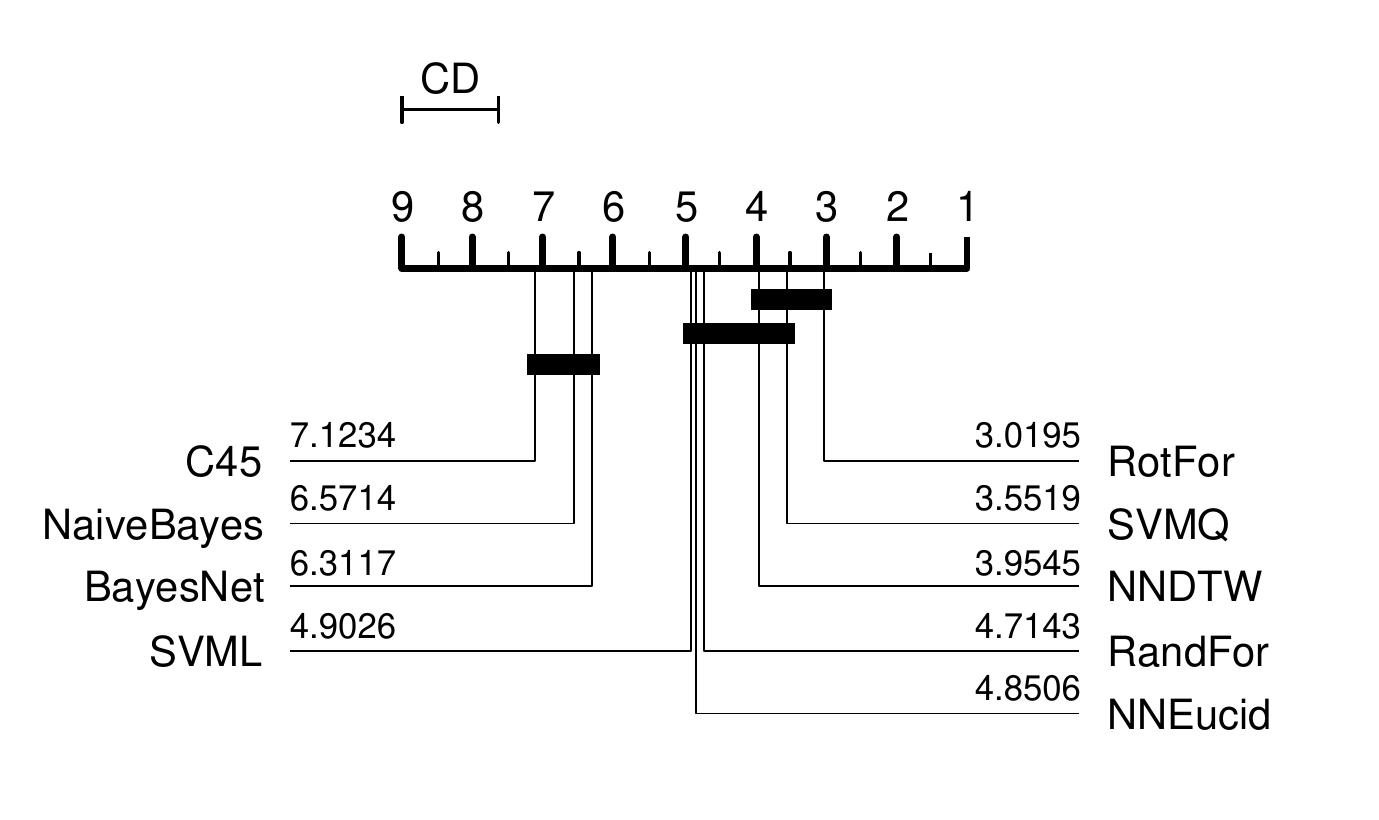}
	\caption{The average ranks for nine different classifiers on 77 data sets. Critical difference is 1.3691}
	\label{nnComp}
\end{figure}

Figure~\ref{nnComp} presents the critical difference diagram for nine different classifiers, all trained with the default Weka parameters. It shows that the bottom group of C4.5, Naive Bayes and the Bayesian Network are significantly worse that the other classifiers. It also shows that there is no significant difference between the top three classifiers (1-NN with DTW, quadratic support vector machine and rotation forest). The middle group of linear SVM, 1-NN Euclidean and Random Forest are tightly grouped, and are all significantly worse than the top performing classifier, rotation forest.

Clearly, there is no evidence to refute the hypothesis concerning 1-NN DTW. Figure~\ref{nnComp} would suggest that 1-NN with Euclidean distance is significantly worse than the rotation forest. Paired tests of difference in average for 1-NN Euclidean vs SVMQ, Rotation Forest and 1-NN DTW all indicate that we should reject the null hypothesis that difference in means and medians is zero in each case. We used a paired T test for the mean and a Wilcoxon signed rank test for the median, full results are available from~\cite{TSC_Web}.  Our conclusion from these experiments is that, whilst it is true that 1-NN DTW is hard to beat, 1-NN with Euclidean distance is beaten by two off-the-shelf classifiers constructed with no parameter optimisation.

Table~\ref{1NNsplit} shows the average ranks of the classifiers split by problem type. This indicates that the NN classifiers perform much better on the motion data sets than the image outline problems. The poor performance on outlines is  caused in part by rotation in the image data sets.

\begin{table}[htbp]
	\centering
  	\caption{Average ranks of 6 of the 9 classifiers split by problem type. Naive Bayes, Bayes Net and C4.5 have been removed for clarity}
  \footnotesize
    \begin{tabular}{|c|c|c|c|c|c|c|} \hline
Type    &NNEucid&NNDTW  &RandF&RotF&SVML&SVMQ\\ \hline
IMAGE   &5.27   &4.52   &4.91&3.18&3.93&3.23\\
MOTION  &3.08   &3.00   &3.71&2.21&6.04&4.13\\
SENSOR  &5.16   &4.23   &4.65&3.08&5.08&3.55\\
SIM     &5.20   &1.40   &6.20&3.90&5.70&4.50\\ \hline
    \end{tabular}
  \label{1NNsplit}
\end{table}

\begin{figure}[htbp]
	\centering
        \includegraphics[width=9cm]{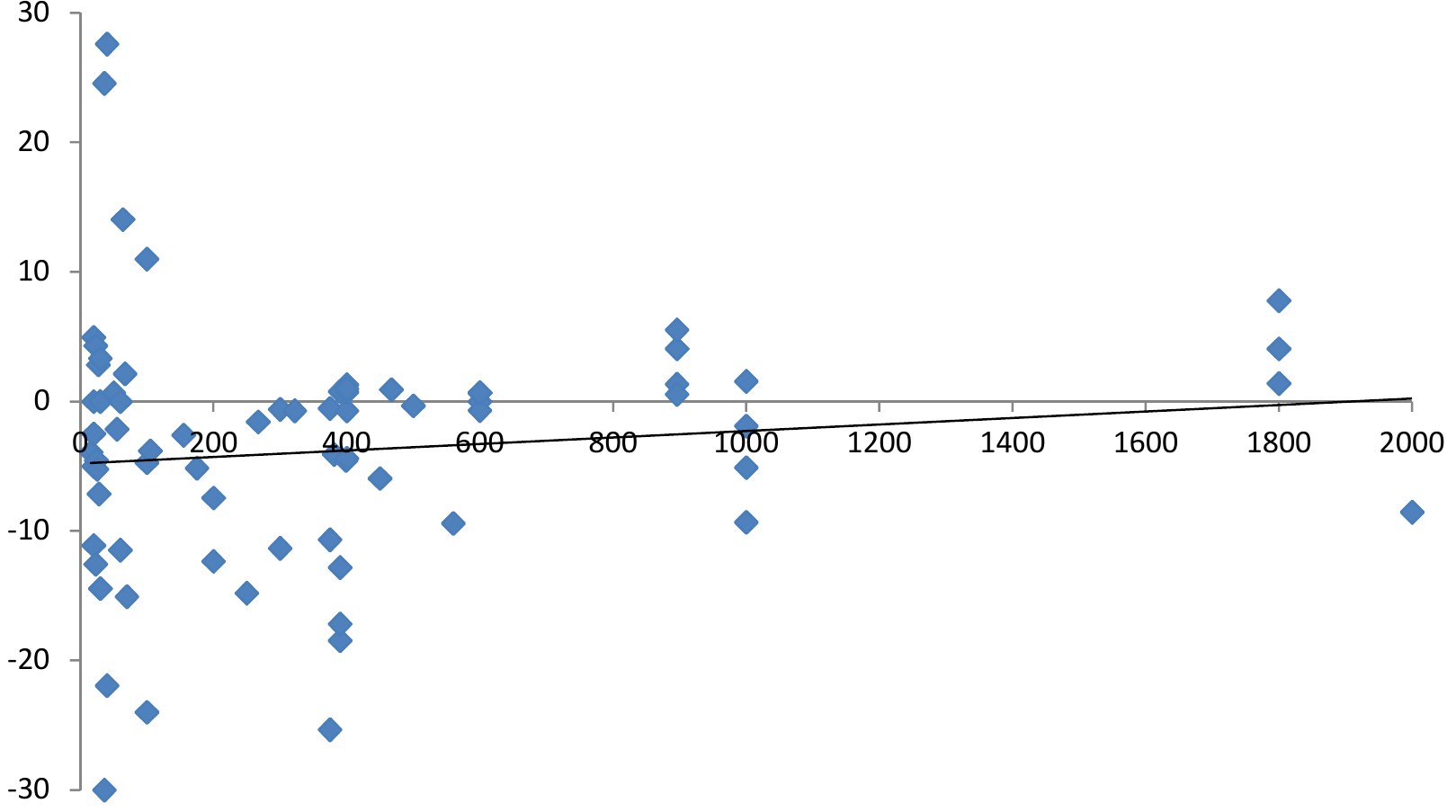}
	\caption{(Euclidean accuracy - DTW accuracy) plotted against the number of training cases and the least squares regression line. The slope of the regression line is not significantly different to zero.}
	\label{fig2}
\end{figure}

We also found no evidence that 1-NN Euclidean converges to 1-NN DTW based on training size, although this is probably because of a lack of problems with large train set sizes. Figure~\ref{fig2} shows the plot of the difference in accuracy of 1-NN Euclidean and 1-NN DTW against the number of training cases.

Our conclusion from these experiments is that 1-NN with Euclidean distance should no longer be used as a standard benchmark against which to compare new algorithms for time series classification. We would recommend that nearest neighbour DTW classifiers should be the new default for distance measure based classifiers, and that the results for SVMQ and rotation forest should also be reported. We address the exact nature of how to use DTW nearest neighbour classifiers in the remainder of this section.

\subsection{Is it worth setting $k$ through cross validation?}

One commonly used method of improving nearest neighbour classifiers is to set $k$ through cross validation on the training data. This requires the calculation of the distance matrix for the training set, and so adds some time overhead. This is particularly time consuming when cross validating against window size for DTW, since every new window size may create a different distance matrix. We are using 7 variants of distance measure with nearest neighbour classifiers: Eulcidean (Euclid), DTW with full warping window (DTWR1), DTW with warping window set through cross validation (DTWRN), Derivative DTW with full and CV set warping windows (DDTWR1 and DDTWRN) and weighted DTW and DDTW (WDTW and WDDTW).

\begin{table}[htbp]
	\centering
  	\caption{Summary statistics on the difference in accuracy between 1-NN and $k$-NN classifiers. The final two columns give the p value for a paired two sample T test for difference in means and
  a Wilcoxon signed rank test for difference in medians}
  \footnotesize
    \begin{tabular}{|c|c|c|c|c|c|c|} \hline
Distance & Mean         &1NN        &Equal  &kNN        &T Test  & Rank \\
Measure  & Difference   & Better    &       & Better    &P value & P Value \\ \hline
Euclidean       &0.16\%&11&48&18&0.3590&0.1748\\
DTWR1           &0.28\%&11&38&28&0.2454&0.0362\\
DTWRN           &1.26\%&18&37&22&0.0004&0.3799\\
DDTWR1    &0.73\%&8&36&33&0.0360&0.0038\\
DDTWRN          &0.17\%&12&42&23&0.3195&0.1044\\
WDTW            &6.67\%&16&36&25&0.0000&0.1989\\
WDDTW     &0.80\%&12&38&27&0.0058&0.0100\\ \hline
    \end{tabular}
  \label{warpingwindow}
\end{table}

Table~\ref{warpingwindow} shows the summary of the improvement in test accuracy of finding $k$ through cross validation rather than setting $k=1$. The largest average improvement is with WDTW, which improved 6.67\%. However, this
was skewed by two data sets with very large improvements, and the non-parametric Wilcoxon signed rank test could not detect a significant difference. The algorithm that showed significant improvement through setting $k$ was Derivative DTW. We conclude that if it is feasible to do so, then it is worthwhile setting $k$ through cross validation, but not doing so is unlikely to have a significant effect on accuracy.

\subsection{Is it worth finding the DTW window size through cross validation?}

\begin{figure*}[htbp]
	\centering
        \includegraphics[width=11cm]{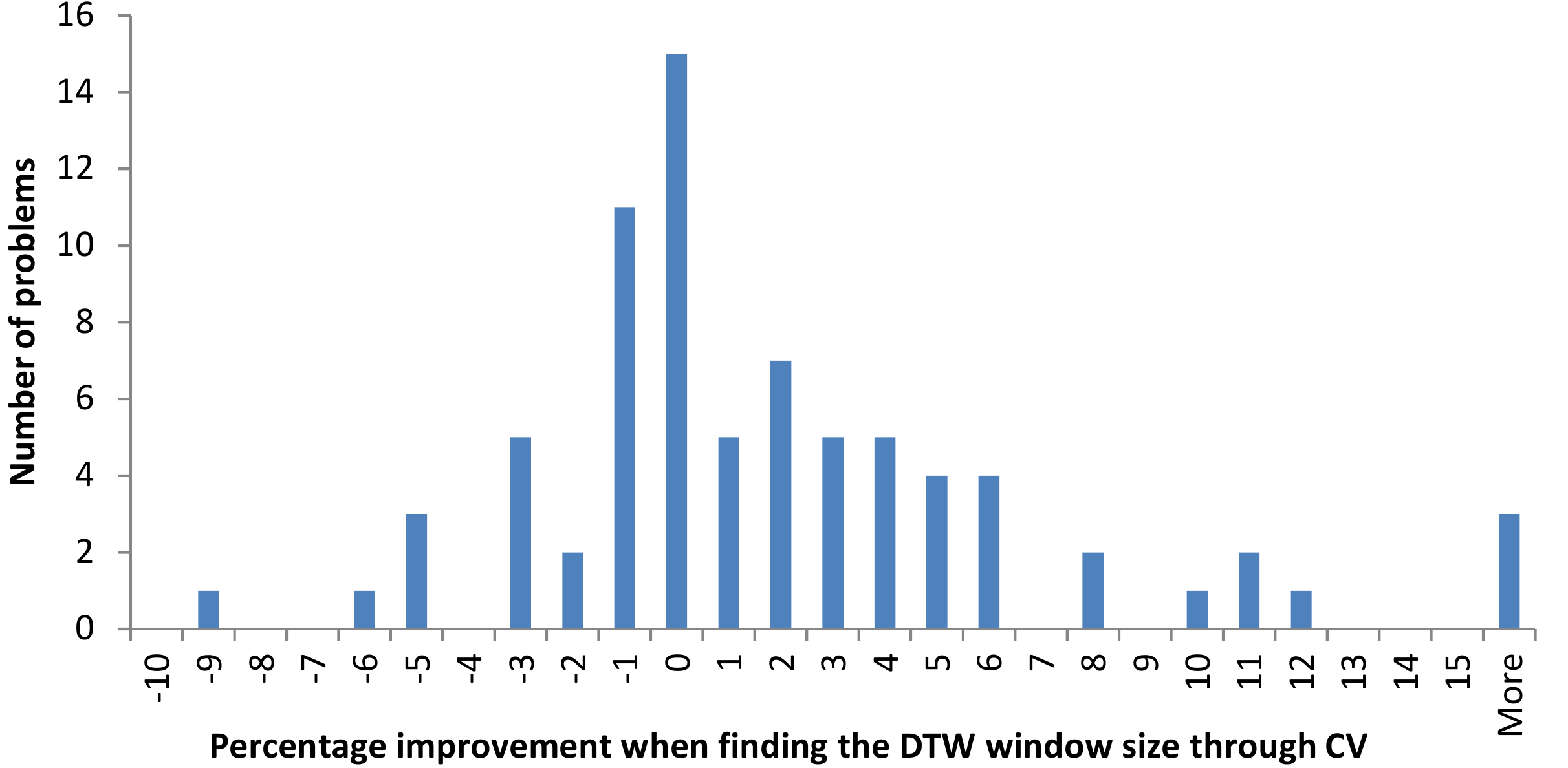}
	\caption{Histogram of the percentage improvement in accuracy when the DTW window size is set through cross validation. }
	\label{dtwcv}
\end{figure*}

The short answer to this question is yes. Figure~\ref{dtwcv} summarises the improvement in test accuracy over all 77 data sets. The mean improvement is 1.8\%  and the median is 0.3\% (both  significant at the 5\% level).

\subsection{Which is better for DTW, setting window size or setting weights?}

\begin{figure}[htbp]
	\centering
        \includegraphics[width=9cm]{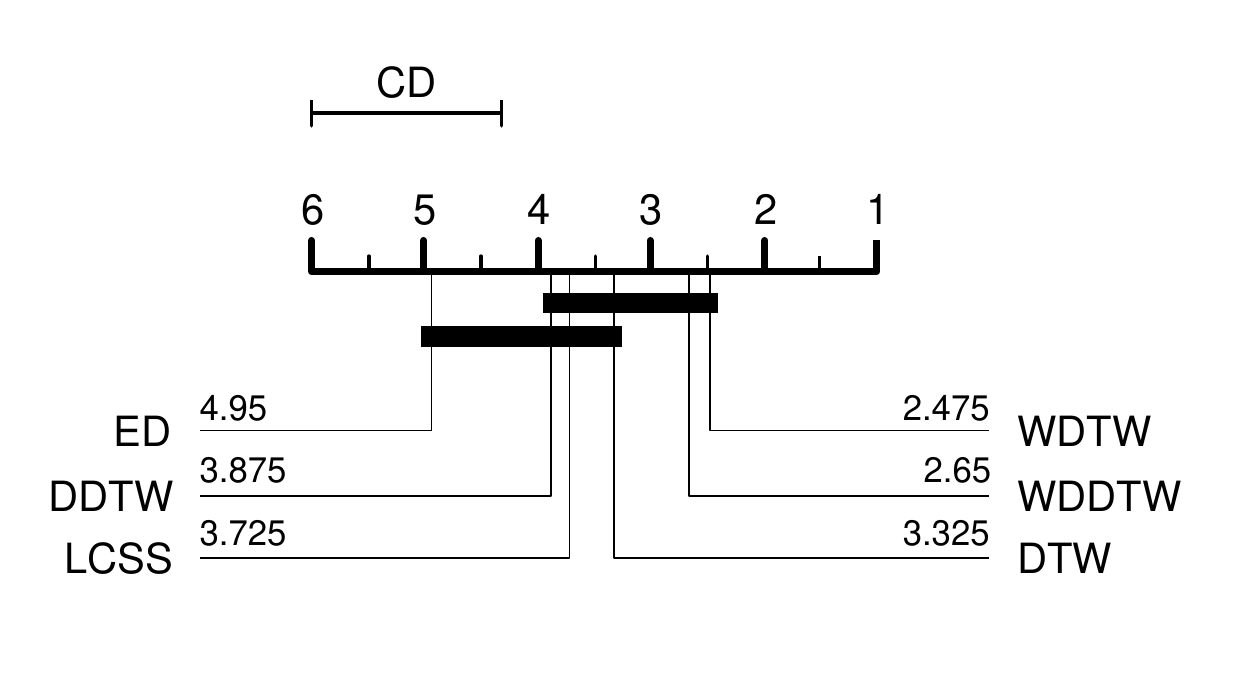}
	\caption{The average ranks for 1-NN classifiers for Euclidean Distance (ED), Longest Common Subsequence (LCSS), Dynamic Time Warping (DTW), Derivative Dynamic Time Warping (DDTW), Dynamic Time Warping with window set through cross validation (DTWCV) and weighted versions (WDTW and WDDTW) for 20 data sets. Data taken directly from~\cite{jeong11weighted}.  }
	\label{cd1}
\end{figure}

The results published for the weighting algorithm proposed in~\cite{jeong11weighted} are summarised in the critical difference diagram in Figure~\ref{cd1}.

Their claim that the weighting leads to clear improvement is not backed up by these results. Figure~\ref{cd1} indicates that although weighted versions have the highest rank, the result cannot be claimed to be significant.
 The only significant difference is between the weighted versions of DTW and Euclidean distance. This demonstrates the need for testing on a large number of data sets. Furthermore, the comparisons they make are biased. Firstly, they compare full window DTW to the weighting scheme, when it would be more appropriate to compare against DTW with window size set through cross validation. Secondly, they use half the testing set data to validate the weighting parameter, but do not allow the other classifiers access to this data set.

The average rank is presented, along with the groups within which there is no significant difference in rank. The average difference in accuracy between DTW and WDTW is just 0.02145, but the difference between DTWCV and WDTW is merely 0.0056. We claim that DTWCV vs WDTW is a fairer comparison because WDTW has had the parameter $g$ set through cross validation.
\begin{figure}[htbp]
	\centering
       \includegraphics[width=9cm]{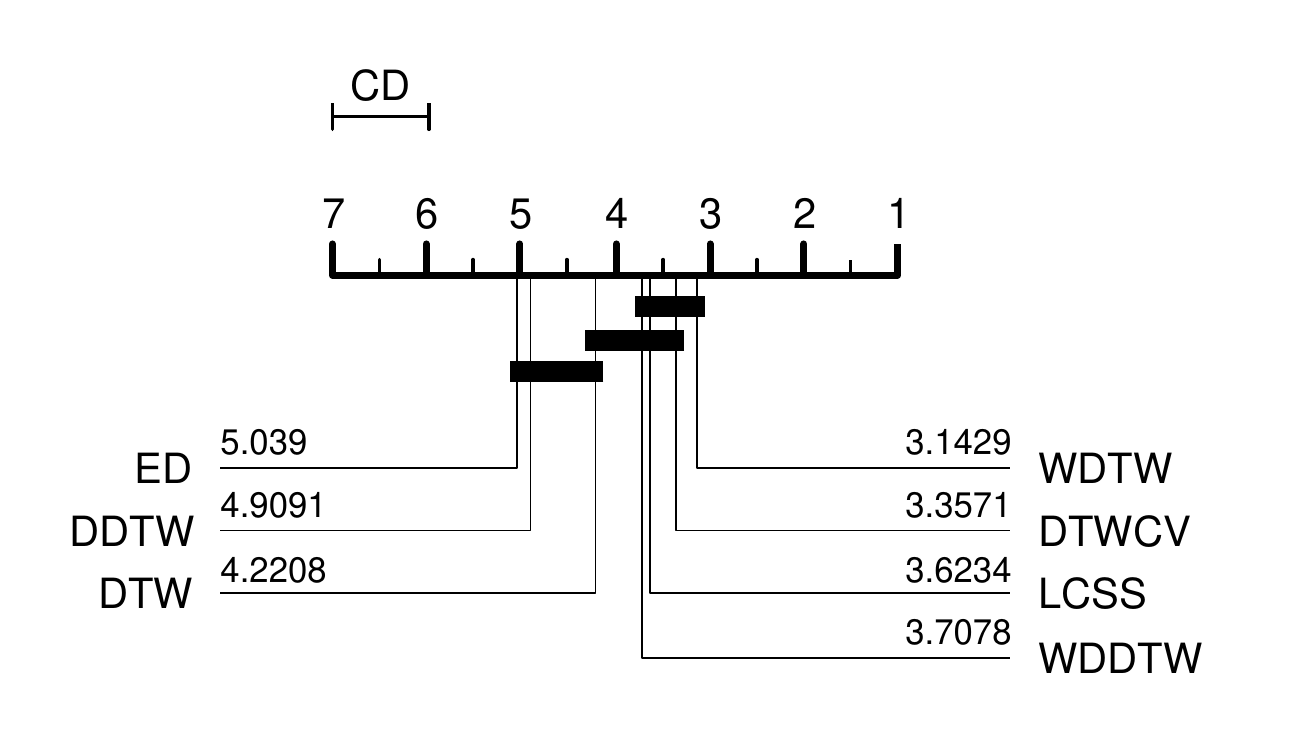}
	\caption{The average ranks for 1-NN classifiers on the 77 data sets described in Section~\ref{data}. The critical difference is 1.0264}
	\label{cd2}
\end{figure}

We have implemented their weighting algorithm and evaluated it on the 77 data sets described in Section~\ref{data}. In our experiments all parameter optimisation is conducted on the training set through cross validation. Figure~\ref{cd2} shows that although weighted DTW still has the highest rank, the difference between WDTW and DTWCV is very small and there is no significant difference between the top four classifiers. There is a significant difference between the full window classifiers and those that restrict the window, but there is no evidence to suggest the weighting algorithm is better than the window algorithm. LCSS performs surprisingly well. Table~\ref{1NN} shows the average ranks split by problem type. LCSS performs well on the motion data and image, but ranks poorly on sensor. Conversely, weighted and windowed DTW rank highly on motion and sensor but relatively poorly on image outlines. LCSS is in many ways closer to a Shapelet approach~\cite{ye09shapelets} than DTW, and this result suggests that subsequence matching techniques such as LCSS and Shapelets may be better for image outline classification.

\begin{table}[htbp]
	\centering
  	\caption{Average ranks of 6 1-NN elastic measure classifiers split by problem type.}
  \footnotesize
    \begin{tabular}{|c|c|c|c|c|c|c|c|} \hline
TYPE    &DTWCV  &DTW&WDTW&DDTW&WDDTW&LCSS\\   \hline
IMAGE   &3.72   &4.60&3.86&4.43&3.19&3.12\\
MOTION  &2.42   &4.25&2.21&6.17&4.79&2.96\\
SENSOR  &3.31   &4.13&2.97&4.84&3.48&4.45\\
SIM     &3.80   &2.50&2.30&5.10&5.50&3.00\\ \hline
 &3.36    &4.22&3.14&4.91&3.71&3.62\\ \hline

 \end{tabular}
  \label{1NN}
\end{table}

\section{Conclusions}
\label{conc}

We have conducted extensive experiments on the largest set of time series classification problems ever used in the literature (to the best of our knowledge). We have performed these tests to validate commonly held assumptions, evaluate a recently published algorithm and assess methods for ensembling.

Firstly, we conclude that comparisons against 1-NN with Euclidean distance for new TSC algorithms are not particularly informative, since three standard classifiers applied to the raw data perform significantly better with no parameter tuning at all. We think that a new algorithm is only of interest in terms of accuracy if it can significantly outperform 1-NN DTW with a full warping window. Secondly, when using a NN classifier with DTW on a new problem, we would advise that it is not particularly important to set $k$ through cross validation, but that setting the warping window size is worthwhile.
Thirdly, we conclude that the weighting algorithm for DTW described in~\cite{jeong11weighted} is significantly better than DTW with a full warping window, but not significantly different to DTW with the window set through cross validation.

\bibliographystyle{IEEEtran}
\bibliography{ICDM2013}

\end{document}